\documentclass{article}

\usepackage{times}
\usepackage{graphicx} 
\usepackage{subfigure} 

\usepackage{natbib}

\usepackage{algorithm}
\usepackage{algorithmic}

\usepackage{amssymb}
\usepackage{booktabs}
\usepackage{amsmath}
\usepackage{etoolbox}
\usepackage{xspace}

\usepackage[T1]{fontenc}

\usepackage[draft=false]{hyperref}

\usepackage[accepted]{icml2017}

\icmltitlerunning{Regularizing and Optimizing LSTM Language Models}

\begin{document}

\newcommand{\ltwo}{$L_2$\xspace}

\twocolumn[
\icmltitle{Regularizing and Optimizing LSTM Language Models}




\begin{icmlauthorlist}
\icmlauthor{Stephen Merity}{sfr}
\icmlauthor{Nitish Shirish Keskar}{sfr}
\icmlauthor{Richard Socher}{sfr}
\end{icmlauthorlist}

\icmlaffiliation{sfr}{Salesforce Research, Palo Alto, USA}
\icmlcorrespondingauthor{Stephen Merity}{smerity@salesforce.com}

\icmlkeywords{machine learning, deep learning, LSTM, regularization, optimization}

\vskip 0.3in
]



\printAffiliationsAndNotice{}  

\begin{abstract} 
Recurrent neural networks (RNNs), such as long short-term memory networks (LSTMs), serve as a fundamental building block for many sequence learning tasks, including machine translation, language modeling, and question answering.
In this paper, we consider the specific problem of word-level language modeling and investigate strategies for regularizing and optimizing LSTM-based models.
We propose the weight-dropped LSTM which uses DropConnect on hidden-to-hidden weights as a form of recurrent regularization.
Further, we introduce NT-ASGD, a variant of the averaged stochastic gradient method, wherein the averaging trigger is determined using a non-monotonic condition as opposed to being tuned by the user.
Using these and other regularization strategies, we achieve state-of-the-art word level perplexities on two data sets: 57.3 on Penn Treebank and 65.8 on WikiText-2.
In exploring the effectiveness of a neural cache in conjunction with our proposed model, we achieve an even lower state-of-the-art perplexity of 52.8 on Penn Treebank and 52.0 on WikiText-2.
\end{abstract}

\section{Introduction}

Effective regularization techniques for deep learning have been the subject of much research in recent years.
Given the over-parameterization of neural networks, generalization performance crucially relies on the ability to regularize the models sufficiently.
Strategies such as dropout \cite{Srivastava2014DropoutAS} and batch normalization \cite{Ioffe2015BatchNA} have found great success and are now ubiquitous in feed-forward and convolutional neural networks.
Na\"ively applying these approaches to the case of recurrent neural networks (RNNs) has not been highly successful however.
Many recent works have hence been focused on the extension of these regularization strategies to RNNs; we briefly discuss some of them below.

A na\"ive application of dropout \cite{Srivastava2014DropoutAS} to an RNN's hidden state is ineffective as it disrupts the RNN's ability to retain long term dependencies \cite{Zaremba2014}.
\citet{Gal2016ATG} propose overcoming this problem by retaining the same dropout mask across multiple time steps as opposed to sampling a new binary mask at each timestep.
Another approach is to regularize the network through limiting updates to the RNN's hidden state.
One such approach is taken by \citet{Semeniuta2016RecurrentDW} wherein the authors drop updates to network units, specifically the input gates of the LSTM, in lieu of the units themselves.
This is reminiscent of zoneout \citep{Krueger2016} where updates to the hidden state may fail to occur for randomly selected neurons.

Instead of operating on the RNN's hidden states, one can regularize the network through restrictions on the recurrent matrices as well.
This can be done either through restricting the capacity of the matrix \citep{Arjovsky2016, Wisdom2016, Jing2016} or through element-wise interactions \citep{Balduzzi2016,Bradbury2016,Seo2016}.

Other forms of regularization explicitly act upon activations such as batch normalization~\citep{Ioffe2015BatchNA}, recurrent batch normalization~\citep{Cooijmans2016RecurrentBN}, and layer normalization~\citep{Ba2016LayerN}.
These all introduce additional training parameters and can complicate the training process while increasing the sensitivity of the model.

In this work, we investigate a set of regularization strategies that are not only highly effective but which can also be used with no modification to existing LSTM implementations.
The weight-dropped LSTM applies recurrent regularization through a DropConnect mask on the hidden-to-hidden recurrent weights.
Other strategies include the use of randomized-length backpropagation through time (BPTT), embedding dropout, activation regularization (AR), and temporal activation regularization (TAR).

As no modifications are required of the LSTM implementation these regularization strategies are compatible with black box libraries, such as NVIDIA cuDNN, which can be many times faster than na\"ive LSTM implementations.

Effective methods for training deep recurrent networks have also been a topic of renewed interest. Once a model has been defined, the training algorithm used is required to not only find a good minimizer of the loss function but also converge to such a minimizer rapidly. The choice of the optimizer is even more important in the context of regularized models since such strategies, especially the use of dropout, can impede the training process. Stochastic gradient descent (SGD), and its variants such as Adam \cite{kingma2014adam} and RMSprop \cite{tieleman2012lecture} are amongst the most popular training methods. These methods iteratively reduce the training loss through scaled (stochastic) gradient steps. In particular, Adam has been found to be widely applicable despite requiring less tuning of its hyperparameters. In the context of word-level language modeling, past work has empirically found that SGD outperforms other methods in not only the final loss but also in the rate of convergence. This is in agreement with recent evidence pointing to the insufficiency of adaptive gradient methods \cite{wilson2017marginal}.

Given the success of SGD, especially within the language modeling domain, we investigate the use of averaged SGD (ASGD)  \cite{polyak1992acceleration} which is known to have superior theoretical guarantees.
ASGD carries out iterations similar to SGD, but instead of returning the last iterate as the solution, returns an average of the iterates past a certain, tuned, threshold $T$.
This threshold $T$ is typically tuned and has a direct impact on the performance of the method.
We propose a variant of ASGD where $T$ is determined on the fly through a non-monotonic criterion and show that it achieves better training outcomes compared to SGD.

\section{Weight-dropped LSTM}

We refer to the mathematical formulation of the LSTM,
\begin{align*}
i_t &= \sigma(W^i x_t + U^i h_{t-1}) \\
f_t &= \sigma(W^f x_t + U^f h_{t-1}) \\
o_t &= \sigma(W^o x_t + U^o h_{t-1}) \\
\tilde{c}_t &= \text{tanh}(W^c x_t + U^c h_{t-1}) \\
c_t &= i_t \odot \tilde{c}_t + f_t \odot + \tilde{c}_{t-1} \\
h_t &= o_t \odot \text{tanh}(c_t)
\end{align*}

where $[W^i, W^f, W^o, U^i, U^f, U^o]$ are weight matrices, $x_t$ is the vector input to the timestep $t$, $h_t$ is the current exposed hidden state, $c_t$ is the memory cell state, and $\odot$ is element-wise multiplication.

Preventing overfitting within the recurrent connections of an RNN has been an area of extensive research in language modeling.
The majority of previous recurrent regularization techniques have acted on the hidden state vector $h_{t-1}$, most frequently introducing a dropout operation between timesteps, or performing dropout on the update to the memory state $c_t$.
These modifications to a standard LSTM prevent the use of black box RNN implementations that may be many times faster due to low-level hardware-specific optimizations.

We propose the use of DropConnect \citep{dropconnect} on the recurrent hidden to hidden weight matrices which does not require any modifications to an RNN's formulation.
As the dropout operation is applied once to the weight matrices, before the forward and backward pass, the impact on training speed is minimal and any standard RNN implementation can be used, including inflexible but highly optimized black box LSTM implementations such as NVIDIA's cuDNN LSTM.

By performing DropConnect on the hidden-to-hidden weight matrices $[U^i, U^f, U^o, U^c]$ within the LSTM, we can prevent overfitting from occurring on the recurrent connections of the LSTM.
This regularization technique would also be applicable to preventing overfitting on the recurrent weight matrices of other RNN cells.

As the same weights are reused over multiple timesteps, the same individual dropped weights remain dropped for the entirety of the forward and backward pass.
The result is similar to variational dropout, which applies the same dropout mask to recurrent connections within the LSTM by performing dropout on $h_{t-1}$, except that the dropout is applied to the recurrent weights.
DropConnect could also be used on the non-recurrent weights of the LSTM $[W^i, W^f, W^o]$ though our focus was on preventing overfitting on the recurrent connection.

\section{Optimization}
SGD is among the most popular methods for training deep learning models across various modalities including computer vision, natural language processing, and reinforcement learning. The training of deep networks can be posed as a non-convex optimization problem
\begin{align*}
\min_w \quad \frac{1}{N} \sum_{i=1}^N f_i(w),
\end{align*}
where $f_i$ is the loss function for the $i^{th}$ data point, $w$ are the weights of the network, and the expectation is taken over the data. Given a sequence of learning rates, $\gamma_k$, SGD iteratively takes steps of the form 
\begin{align}
w_{k+1} = w_k - \gamma_k \hat{\nabla} f(w_k), \label{eq:sgd}
\end{align}
where the subscript denotes the iteration number and the $\hat{\nabla}$ denotes a stochastic gradient that may be computed on a minibatch of data points. SGD demonstrably performs well in practice and also possesses several attractive theoretical properties such as linear convergence \cite{bottou2016optimization}, saddle point avoidance \cite{panageas2016gradient} and better generalization performance \cite{hardt2015train}.
For the specific task of neural language modeling, traditionally SGD without momentum has been found to outperform other algorithms such as momentum SGD \cite{sutskever2013importance}, Adam \cite{kingma2014adam}, Adagrad \cite{duchi2011adaptive} and RMSProp \cite{tieleman2012lecture} by a statistically significant margin.

Motivated by this observation, we investigate averaged SGD (ASGD) to further improve the training process. ASGD has been analyzed in depth theoretically and many surprising results have been shown including its asymptotic second-order convergence \cite{polyak1992acceleration,mandt2017stochastic}. ASGD takes steps identical to equation \eqref{eq:sgd} but instead of returning the last iterate as the solution, returns $\frac{1}{(K-T+1)} \sum_{i=T}^K w_i$, where $K$ is the total number of iterations and $T<K$ is a user-specified averaging trigger. 

\begin{algorithm}
\caption{Non-monotonically Triggered ASGD (NT-ASGD)}
\label{alg:asgd}
{\bf Inputs:} Initial point $w_0$, learning rate $\gamma$, logging interval $L$, non-monotone interval $n$.
\begin{algorithmic}[1]
\STATE Initialize $k\gets 0$, $t\gets 0$, $T \gets 0$, \texttt{logs} $\gets $ \texttt{[]}
\WHILE{stopping criterion not met}
\STATE Compute stochastic gradient $\hat{\nabla} f(w_k)$ and take SGD step \eqref{eq:sgd}.
\IF{ \texttt{mod}$(k,L)=0$ and $T=0$}
\STATE Compute validation perplexity $v$.
\IF{$t>n$ and $ \displaystyle v > \min_{l \in \{t-n,\cdots,t\}}$ \texttt{logs[l]}}
\STATE Set $T \gets k$
\ENDIF
\STATE Append $v$ to \texttt{logs}
\STATE $t \gets t + 1$
\ENDIF
\ENDWHILE
\end{algorithmic}
{\bf return }  $\frac{\sum_{i=T}^k w_i}{(k-T+1)}$
\end{algorithm}

Despite its theoretical appeal, ASGD has found limited practical use in training of deep networks.
This may be in part due to unclear tuning guidelines for the learning-rate schedule $\gamma_k$ and averaging trigger $T$.
If the averaging is triggered too soon, the efficacy of the method is impacted, and if it is triggered too late, many additional iterations may be needed to converge to the solution.
In this section, we describe a non-monotonically triggered variant of ASGD (NT-ASGD), which obviates the need for tuning $T$. Further, the algorithm uses a constant learning rate throughout the experiment and hence no further tuning is necessary for the decay scheduling.

Ideally, averaging needs to be triggered when the SGD iterates converge to a steady-state distribution \cite{mandt2017stochastic}. This is roughly equivalent to the convergence of SGD to a neighborhood around a solution. In the case of SGD, certain learning-rate reduction strategies such as the step-wise strategy analogously reduce the learning rate by a fixed quantity at such a point. A common strategy employed in language modeling is to reduce the learning rates by a fixed proportion when the performance of the model's primary metric (such as perplexity) worsens or stagnates. Along the same lines, one could make a triggering decision based on the performance of the model on the validation set. However, instead of averaging immediately after the validation metric worsens, we propose a non-monotonic criterion that conservatively triggers the averaging when the validation metric fails to improve for multiple cycles; see Algorithm \ref{alg:asgd}. Given that the choice of triggering is irreversible, this conservatism ensures that the randomness of training does not play a major role in the decision. Analogous strategies have also been proposed for learning-rate reduction in SGD \cite{keskar2015nonmonotone}.

While the algorithm introduces two additional hyperparameters, the logging interval $L$ and non-monotone interval $n$, we found that setting $L$ to be the number of iterations in an epoch and $n=5$ worked well across various models and data sets. As such, we use this setting in all of our NT-ASGD experiments in the following section and demonstrate that it achieves better training outcomes as compared to SGD.

\section{Extended regularization techniques}

In addition to the regularization and optimization techniques above, we explored additional regularization techniques that aimed to improve data efficiency during training and to prevent overfitting of the RNN model.

\subsection{Variable length backpropagation sequences}

Given a fixed sequence length that is used to break a data set into fixed length batches, the data set is not efficiently used.
To illustrate this, imagine being given $100$ elements to perform backpropagation through with a fixed backpropagation through time (BPTT) window of $10$.
Any element divisible by $10$ will never have any elements to backprop into, no matter how many times you may traverse the data set.
Indeed, the backpropagation window that each element receives is equal to $i \bmod 10$ where $i$ is the element's index.
This is data inefficient, preventing $\frac{1}{10}$ of the data set from ever being able to improve itself in a recurrent fashion, and resulting in $\frac{8}{10}$ of the remaining elements receiving only a partial backpropagation window compared to the full possible backpropagation window of length $10$.

To prevent such inefficient data usage, we randomly select the sequence length for the forward and backward pass in two steps.
First, we select the base sequence length to be $\text{seq}$ with probability $p$ and $\frac{\text{seq}}{2}$ with probability $1-p$, where $p$ is a high value approaching $1$.
This spreads the starting point for the BPTT window beyond the base sequence length.
We then select the sequence length according to $\mathcal{N}(\text{seq}, s)$, where $\text{seq}$ is the base sequence length and $s$ is the standard deviation.
This jitters the starting point such that it doesn't always fall on a specific word divisible by $\text{seq}$ or $\frac{\text{seq}}{2}$.
From these, the sequence length more efficiently uses the data set, ensuring that when given enough epochs all the elements in the data set experience a full BPTT window, while ensuring the average sequence length remains around the base sequence length for computational efficiency.

During training, we rescale the learning rate depending on the length of the resulting sequence compared to the original specified sequence length.
The rescaling step is necessary as sampling arbitrary sequence lengths with a fixed learning rate favors short sequences over longer ones.
This linear scaling rule has been noted as important for training large scale minibatch SGD without loss of accuracy \citep{Goyal2017} and is a component of unbiased truncated backpropagation through time \citep{Tallec2017}.

\subsection{Variational dropout}
In standard dropout, a new binary dropout mask is sampled each and every time the dropout function is called.
New dropout masks are sampled even if the given connection is repeated, such as the input $x_0$ to an LSTM at timestep $t=0$ receiving a different dropout mask than the input $x_1$ fed to the same LSTM at $t=1$.
A variant of this, variational dropout \cite{Gal2016ATG}, samples a binary dropout mask only once upon the first call and then to repeatedly use that locked dropout mask for all repeated connections within the forward and backward pass.

While we propose using DropConnect rather than variational dropout to regularize the hidden-to-hidden transition within an RNN, we use variational dropout for all other dropout operations, specifically using the same dropout mask for all inputs and outputs of the LSTM within a given forward and backward pass.
Each example within the minibatch uses a unique dropout mask, rather than a single dropout mask being used over all examples, ensuring diversity in the elements dropped out.

\subsection{Embedding dropout}

Following \citet{Gal2016ATG}, we employ embedding dropout.
This is equivalent to performing dropout on the embedding matrix at a word level, where the dropout is broadcast across all the word vector's embedding.
The remaining non-dropped-out word embeddings are scaled by $\frac{1}{1 - p_e}$ where $p_e$ is the probability of embedding dropout.
As the dropout occurs on the embedding matrix that is used for a full forward and backward pass, this means that all occurrences of a specific word will disappear within that pass, equivalent to performing variational dropout on the connection between the one-hot embedding and the embedding lookup.

\subsection{Weight tying}

Weight tying \citep{Inan2016, Press2016} shares the weights between the embedding and softmax layer, substantially reducing the total parameter count in the model.
The technique has theoretical motivation \citep{Inan2016} and prevents the model from having to learn a one-to-one correspondence between the input and output, resulting in substantial improvements to the standard LSTM language model.

\subsection{Independent embedding size and hidden size}

In most natural language processing tasks, both pre-trained and trained word vectors are of relatively low dimensionality---frequently between 100 and 400 dimensions in size.
Most previous LSTM language models tie the dimensionality of the word vectors to the dimensionality of the LSTM's hidden state.
Even if reducing the word embedding size was not beneficial in preventing overfitting, the easiest reduction in total parameters for a language model is reducing the word vector size.
To achieve this, the first and last LSTM layers are modified such that their input and output dimensionality respectively are equal to the reduced embedding size.

\subsection{Activation Regularization (AR) and Temporal Activation Regularization (TAR)}
\ltwo-regularization is often used on the weights of the network to control the norm of the resulting model and reduce overfitting. In addition, \ltwo decay can be used on the individual unit activations and on the difference in outputs of an RNN at different time steps; these strategies labeled as activation regularization (AR) and temporal activation regularization (TAR) respectively \cite{smerity-revisiting}.
AR penalizes activations that are significantly larger than $0$ as a means of regularizing the network. Concretely, AR is defined as
\[
\alpha \, L_2 (m \odot h_t) 
\]
where $m$ is the dropout mask, $L_2(\cdot) = \lVert \cdot \rVert_2$, $h_t$ is the output of the RNN at timestep $t$, and $\alpha$ is a scaling coefficient. TAR falls under the broad category of \textit{slowness} regularizers \citep{Hinton1989, Foldiak1991, Luciw2012, Jonschkowski2015LearningSR} which penalize the model from producing large changes in the hidden state. Using the notation from AR, TAR is defined as
\[
\beta \, L_2 (h_t - h_{t+1})
\]
where $\beta$ is a scaling coefficient.
As in \citet{smerity-revisiting}, the AR and TAR loss are only applied to the output of the final RNN layer as opposed to being applied to all layers.

\section{Experiment Details}

For evaluating the impact of these approaches, we perform language modeling over a preprocessed version of the Penn Treebank (PTB) \citep{Mikolov2010} and the WikiText-2 (WT2) data set \citep{Merity2016}.

\textbf{PTB:} The Penn Treebank data set has long been a central data set for experimenting with language modeling.
The data set is heavily preprocessed and does not contain capital letters, numbers, or punctuation.
The vocabulary is also capped at 10,000 unique words, quite small in comparison to most modern datasets, which results in a large number of out of vocabulary (OoV) tokens. 

\textbf{WT2:} WikiText-2 is sourced from curated Wikipedia articles and is approximately twice the size of the PTB data set.
The text is tokenized and processed using the Moses tokenizer \citep{Koehn2007MosesOS}, frequently used for machine translation, and features a vocabulary of over 30,000 words.
Capitalization, punctuation, and numbers are retained in this data set.

All experiments use a three-layer LSTM model with $1150$ units in the hidden layer and
an embedding of size $400$. The loss was averaged over all examples and timesteps.
All embedding weights were uniformly initialized in the interval $[-0.1, 0.1]$ and all other weights were initialized between $[-\frac{1}{\sqrt{H}}, \frac{1}{\sqrt{H}}]$, where $H$ is the hidden size.

For training the models, we use the NT-ASGD algorithm discussed in the previous section for $750$ epochs with $L$ equivalent to one epoch and $n=5$. We use a batch size of $80$ for WT2 and $40$ for PTB. Empirically, we found relatively large batch sizes (e.g., $40$-$80$) performed better than smaller sizes (e.g., $10$-$20$) for NT-ASGD. After completion, we run ASGD with $T=0$ and hot-started $w_0$ as a fine-tuning step to further improve the solution. For this fine-tuning step, we terminate the run using the same non-monotonic criterion detailed in Algorithm 1.

We carry out gradient clipping with maximum norm $0.25$ and use an initial learning rate of $30$ for all experiments.
We use a random BPTT length which is $\mathcal{N}(70, 5)$ with probability $0.95$ and $\mathcal{N}(35, 5)$ with probability $0.05$.
The values used for dropout on the word vectors, the output between LSTM layers, the output of the final LSTM layer, and embedding dropout where $(0.4, 0.3, 0.4, 0.1)$ respectively.
For the weight-dropped LSTM, a dropout of $0.5$ was applied to the recurrent weight matrices.
For WT2, we increase the input dropout to $0.65$ to account for the increased vocabulary size.
For all experiments, we use AR and TAR values of $2$ and $1$ respectively, and tie the embedding and softmax weights.
These hyperparameters were chosen through trial and error and we expect further improvements may be possible if a fine-grained hyperparameter search were to be conducted.
In the results, we abbreviate our approach as AWD-LSTM for ASGD Weight-Dropped LSTM.

\begin{table*}
\center
\begin{tabular}{l|ccc}
\toprule
\bf Model & \bf Parameters & \bf Validation &  \bf Test \\
\midrule
\citet{Mikolov2012} - KN-5 & 2M$^\ddagger$ & $-$ & $141.2$ \\
\citet{Mikolov2012} - KN5 + cache & 2M$^\ddagger$ & $-$ & $125.7$ \\
\citet{Mikolov2012} - RNN & 6M$^\ddagger$ & $-$ & $124.7$ \\
\citet{Mikolov2012} - RNN-LDA & 7M$^\ddagger$ & $-$ & $113.7$ \\
\citet{Mikolov2012} - RNN-LDA + KN-5 + cache & 9M$^\ddagger$ & $-$ & $92.0$ \\
\citet{Zaremba2014} - LSTM (medium) & 20M & $86.2$ & $82.7$ \\
\citet{Zaremba2014} - LSTM (large) & 66M & $82.2$ & $78.4$ \\
\citet{Gal2016ATG} - Variational LSTM (medium) & 20M & $81.9 \pm 0.2$ & $79.7 \pm 0.1$ \\
\citet{Gal2016ATG} - Variational LSTM (medium, MC) & 20M & $-$ & $78.6 \pm 0.1$ \\
\citet{Gal2016ATG} - Variational LSTM (large) & 66M & $77.9 \pm 0.3$ & $75.2 \pm 0.2$ \\
\citet{Gal2016ATG} - Variational LSTM (large, MC) & 66M & $-$ & $73.4 \pm 0.0$ \\
\citet{Kim2016} - CharCNN & 19M & $-$ & $78.9$ \\
\citet{Merity2016} - Pointer Sentinel-LSTM & 21M & $72.4$ & $70.9$ \\
\citet{Grave2016} - LSTM & $-$ & $-$ & $82.3$ \\
\citet{Grave2016} - LSTM + continuous cache pointer & $-$ & $-$ & $72.1$ \\
\citet{Inan2016} - Variational LSTM (tied) + augmented loss & 24M & $75.7$ & $73.2$ \\
\citet{Inan2016} - Variational LSTM (tied) + augmented loss & 51M & $71.1$ & $68.5$ \\
\citet{Zilly2016} - Variational RHN (tied) & 23M & $67.9$ & $65.4$ \\
\citet{Zoph2016} - NAS Cell (tied) & 25M & $-$ & $64.0$ \\
\citet{Zoph2016} - NAS Cell (tied) & 54M & $-$ & $62.4$ \\
\citet{Melis2017} - 4-layer skip connection LSTM (tied) & 24M & $60.9$ & $58.3$ \\
\midrule
AWD-LSTM - 3-layer LSTM (tied) & 24M & $60.0$ & $57.3$ \\
\midrule
AWD-LSTM - 3-layer LSTM (tied) + continuous cache pointer & 24M & $53.9$ & $52.8$ \\
\bottomrule
\end{tabular}
\caption{
Single model perplexity on validation and test sets for the Penn Treebank language modeling task.
Parameter numbers with $\ddagger$ are estimates based upon our understanding of the model and with reference to \citet{Merity2016}.
Models noting \textit{tied} use weight tying on the embedding and softmax weights.
Our model, AWD-LSTM, stands for ASGD Weight-Dropped LSTM.
}
\label{table:PTBresults}
\end{table*}

\begin{table*}
\center
\begin{tabular}{l|ccc}
\toprule
\bf Model & \bf Parameters & \bf Validation &  \bf Test \\
\midrule
\citet{Inan2016} - Variational LSTM  (tied) ($h=650$) & 28M & $92.3$ & $87.7$ \\
\citet{Inan2016} - Variational LSTM  (tied) ($h=650$) + augmented loss & 28M & $91.5$ & $87.0$ \\
\citet{Grave2016} - LSTM & $-$ & $-$ & $99.3$ \\
\citet{Grave2016} - LSTM + continuous cache pointer & $-$ & $-$ & $68.9$ \\
\citet{Melis2017} - 1-layer LSTM (tied) & 24M & $69.3$ & $65.9$ \\
\citet{Melis2017} - 2-layer skip connection LSTM (tied) & 24M & $69.1$ & $65.9$ \\
\midrule
AWD-LSTM - 3-layer LSTM (tied) & 33M & $68.6$ & $65.8$ \\
\midrule
AWD-LSTM - 3-layer LSTM (tied) + continuous cache pointer & 33M & $53.8$ & $52.0$ \\
\bottomrule
\end{tabular}
\caption{Single model perplexity over WikiText-2.
Models noting \textit{tied} use weight tying on the embedding and softmax weights.
Our model, AWD-LSTM, stands for ASGD Weight-Dropped LSTM.
}
\label{table:SelfWT2results}
\end{table*}

\section{Experimental Analysis}

We present the single-model perplexity results for both our models (AWD-LSTM) and other competitive models in Table \ref{table:PTBresults} and \ref{table:SelfWT2results} for PTB and WT2 respectively.
On both data sets we improve the state-of-the-art, with our vanilla LSTM model beating the state of the art by approximately 1 unit on PTB and 0.1 units on WT2.

In comparison to other recent state-of-the-art models, our model uses a vanilla LSTM.
\citet{Zilly2016} propose the recurrent highway network, which extends the LSTM to allow multiple hidden state updates per timestep.
\citet{Zoph2016} use a reinforcement learning agent to generate an RNN cell tailored to the specific task of language modeling, with the cell far more complex than the LSTM.

Independently of our work, \citet{Melis2017} apply extensive hyperparameter search to an LSTM based language modeling implementation, analyzing the sensitivity of RNN based language models to hyperparameters.
Unlike our work, they use a modified LSTM, which caps the input gate $i_t$ to be $\min(1 - f_t, i_t)$, use Adam with $\beta_1=0$ rather than SGD or ASGD, use skip connections between LSTM layers, and use a black box hyperparameter tuner for exploring models and settings.
Of particular interest is that their hyperparameters were tuned individually for each data set compared to our work which shared almost all hyperparameters between PTB and WT2, including the embedding and hidden size for both data sets.
Due to this, they used less model parameters than our model and found shallow LSTMs of one or two layers worked best for WT2.

Like our work, \citet{Melis2017} find that the underlying LSTM architecture can be highly effective compared to complex custom architectures when well tuned hyperparameters are used.
The approaches used in our work and \citet{Melis2017} may be complementary and would be worth exploration.

\section{Pointer models}

In past work, pointer based attention models have been shown to be highly effective in improving language modeling \citep{Merity2016,Grave2016}.
Given such substantial improvements to the underlying neural language model, it remained an open question as to how effective pointer augmentation may be, especially when improvements such as weight tying may act in mutually exclusive ways.

The neural cache model \citep{Grave2016} can be added on top of a pre-trained language model at negligible cost.
The neural cache stores the previous hidden states in memory cells and then uses a simple convex combination of the probability distributions suggested by the cache and the language model for prediction.
The cache model has three hyperparameters: the memory size (window) for the cache, the coefficient of the combination (which determines how the two distributions are mixed), and the flatness of the cache distribution.
All of these are tuned on the validation set once a trained language model has been obtained and require no training by themselves, making it quite inexpensive to use. The tuned values for these hyperparameters were $(2000, 0.1, 1.0)$ for PTB and $(3785, 0.1279, 0.662)$ for WT2 respectively.

In Tables 1 and 2, we show that the model further improves the perplexity of the language model by as much as $6$ perplexity points for PTB and $11$ points for WT2.
While this is smaller than the gains reported in \citet{Grave2016}, which used an LSTM without weight tying, this is still a substantial drop.
Given the simplicity of the neural cache model, and the lack of any trained components, these results suggest that existing neural language models remain fundamentally lacking, failing to capture long term dependencies or remember recently seen words effectively.

To understand the impact the pointer had on the model, specifically the validation set perplexity, we detail the contribution that each word has on the cache model's overall perplexity in Table 3.
We compute the sum of the total difference in the loss function value (i.e., log perplexity) between the LSTM-only and LSTM-with-cache models for the target words in the validation portion of the WikiText-2 data set.
We present results for the sum of the difference as opposed to the mean since the latter undesirably overemphasizes infrequently occurring words for which the cache helps significantly and ignores frequently occurring words for which the cache provides modest improvements that cumulatively make a strong contribution.

The largest cumulative gain is in improving the handling of <unk> tokens, though this is over 11540 instances.
The second best improvement, approximately one fifth the gain given by the <unk> tokens, is for Meridian, yet this word only occurs 161 times.
This indicates the cache still helps significantly even for relatively rare words, further demonstrated by Churchill, Blythe, or Sonic.
The cache is not beneficial when handling frequent word categories, such as punctuation or stop words, for which the language model is likely well suited.
These observations motivate the design of a cache framework that is more aware of the relative strengths of the two models.

\begin{table}
\begin{center}\small
\begin{tabular}{lrr|lrr}
\toprule[0.3ex]
Word & Count & $\Delta$loss & Word & Count & $\Delta$loss \\
 
\midrule[0.2ex]
. & 7632 & -696.45 & <unk> & 11540 & 5047.34 \\
, & 9857 & -687.49 & Meridian & 161 & 1057.78 \\
of & 5816 & -365.21 & Churchill & 137 & 849.43 \\
= & 2884 & -342.01 & - & 67 & 682.15 \\
to & 4048 & -283.10 & Blythe & 97 & 554.95 \\
in & 4178 & -222.94 & Sonic & 75 & 543.85 \\
<eos> & 3690 & -216.42 & Richmond & 101 & 429.18 \\
and & 5251 & -215.38 & Starr & 74 & 416.52 \\
the & 12481 & -209.97 & Australian & 234 & 366.36 \\
a & 3381 & -149.78 & Pagan & 54 & 365.19 \\
" & 2540 & -127.99 & Asahi & 39 & 316.24 \\
that & 1365 & -118.09 & Japanese & 181 & 295.97 \\
by & 1252 & -113.05 & Hu & 43 & 285.58 \\
was & 2279 & -107.95 & Hedgehog & 29 & 266.48 \\
) & 1101 & -94.74 & Burma & 35 & 263.65 \\
with & 1176 & -93.01 & 29 & 92 & 260.88 \\
for & 1215 & -87.68 & Mississippi & 72 & 241.59 \\
on & 1485 & -81.55 & German & 108 & 241.23 \\
as & 1338 & -77.05 & mill & 67 & 237.76 \\
at & 879 & -59.86 & Cooke & 33 & 231.11 \\
\bottomrule[0.3ex]
\end{tabular}
\end{center}
\caption{
The sum total difference in loss (log perplexity) that a given word results in over all instances in the validation data set of WikiText-2 when the continuous cache pointer is introduced. The right column contains the words with the twenty best improvements (i.e., where the cache was advantageous), and the left column the twenty most deteriorated (i.e., where the cache was disadvantageous).}
\label{tb:ptranalysis}
\end{table}

\section{Model Ablation Analysis}

\begin{table*}
\center
\begin{tabular}{l|cc|cc}
\toprule
 & \multicolumn{2}{c|}{PTB} & \multicolumn{2}{c}{WT2} \\
\bf Model & \bf Validation &  \bf Test & \bf Validation &  \bf Test \\
\midrule

AWD-LSTM (tied) & $60.0$ & $57.3$ & 68.6 & 65.8\\
\midrule
-- fine-tuning & $60.7$ & $58.8$ & $69.1$ & $66.0$ \\
-- NT-ASGD & 66.3 & 63.7 & 73.3 & 69.7 \\
\midrule
-- variable sequence lengths & 61.3 & 58.9 & 69.3 & 66.2 \\
-- embedding dropout & 65.1 & 62.7 & 71.1 & 68.1 \\
-- weight decay & 63.7 & 61.0 & 71.9 & 68.7 \\
-- AR/TAR & 62.7 & 60.3 & 73.2 & 70.1 \\ 
-- full sized embedding & 68.0 & 65.6 & 73.7 & 70.7 \\
-- weight-dropping & 71.1 & 68.9 & 78.4 & 74.9 \\
\bottomrule
\end{tabular}
\caption{
Model ablations for our best LSTM models reporting results over the validation and test set on Penn Treebank and WikiText-2.
Ablations are split into optimization and regularization variants, sorted according to the achieved validation perplexity on WikiText-2.
}
\label{table:ablation-ptb}
\end{table*}

In Table \ref{table:ablation-ptb}, we present the values of validation and testing perplexity for different variants of our best-performing LSTM model.
Each variant removes a form of optimization or regularization.

The first two variants deal with the optimization of the language models while the rest deal with the regularization.
For the model using SGD with learning rate reduced by 2 using the same nonmonotonic fashion, there is a significant degradation in performance.
This stands as empirical evidence regarding the benefit of averaging of the iterates.
Using a monotonic criterion instead also hampered performance.
Similarly, the removal of the fine-tuning step expectedly also degrades the performance. This step helps improve the estimate of the minimizer by resetting the memory of the previous experiment. While this process of fine-tuning can be repeated multiple times, we found little benefit in repeating it more than once.

The removal of regularization strategies paints a similar picture; the inclusion of all of the proposed strategies was pivotal in ensuring state-of-the-art performance.
The most extreme perplexity jump was in removing the hidden-to-hidden LSTM regularization provided by the weight-dropped LSTM.
Without such hidden-to-hidden regularization, perplexity rises substantially, up to 11 points.
This is in line with previous work showing the necessity of recurrent regularization in state-of-the-art models \citep{Gal2016ATG,Inan2016}.

We also experiment with static sequence lengths which we had hypothesized would lead to inefficient data usage.
This also worsens the performance by approximately one perplexity unit. Next, we experiment with reverting to matching the sizes of the embedding vectors and the hidden states.
This significantly increases the number of parameters in the network (to $43$M in the case of PTB and $70$M for WT2) and leads to degradation by almost $8$ perplexity points, which we attribute to overfitting in the word embeddings.
While this could potentially be improved with more aggressive regularization, the computational overhead involved with substantially larger embeddings likely outweighs any advantages.
Finally, we experiment with the removal of embedding dropout, AR/TAR and weight decay. In all of the cases, the model suffers a perplexity increase of $2$--$6$ points which we hypothesize is due to insufficient regularization in the network.

\section{Conclusion}

In this work, we discuss regularization and optimization strategies for neural language models. We propose the weight-dropped LSTM, a strategy that uses a DropConnect mask on the hidden-to-hidden weight matrices, as a means to prevent overfitting across the recurrent connections. Further, we investigate the use of averaged SGD with a non-monontonic trigger for training language models and show that it outperforms SGD by a significant margin. We investigate other regularization strategies including the use of variable BPTT length and achieve a new state-of-the-art perplexity on the PTB and WikiText-2 data sets. Our models outperform custom-built RNN cells and complex regularization strategies that preclude the possibility of using optimized libraries such as the NVIDIA cuDNN LSTM. Finally, we explore the use of a neural cache in conjunction with our proposed model and show that this further improves the performance, thus attaining an even lower state-of-the-art perplexity.
While the regularization and optimization strategies proposed are demonstrated on the task of language modeling, we anticipate that they would be generally applicable across other sequence learning tasks.





\bibliography{references}

\begin{thebibliography}{44}
\providecommand{\natexlab}[1]{#1}
\providecommand{\url}[1]{\texttt{#1}}
\expandafter\ifx\csname urlstyle\endcsname\relax
  \providecommand{\doi}[1]{doi: #1}\else
  \providecommand{\doi}{doi: \begingroup \urlstyle{rm}\Url}\fi

\bibitem[Arjovsky et~al.(2016)Arjovsky, Shah, and Bengio]{Arjovsky2016}
Arjovsky, M., Shah, A., and Bengio, Y.
\newblock Unitary evolution recurrent neural networks.
\newblock In \emph{International Conference on Machine Learning}, pp.\
  1120--1128, 2016.

\bibitem[Ba et~al.(2016)Ba, Kiros, and Hinton]{Ba2016LayerN}
Ba, J., Kiros, J., and Hinton, G.~E.
\newblock Layer normalization.
\newblock \emph{CoRR}, abs/1607.06450, 2016.

\bibitem[Balduzzi \& Ghifary(2016)Balduzzi and Ghifary]{Balduzzi2016}
Balduzzi, D. and Ghifary, M.
\newblock Strongly-typed recurrent neural networks.
\newblock \emph{arXiv preprint arXiv:1602.02218}, 2016.

\bibitem[Bottou et~al.(2016)Bottou, Curtis, and
  Nocedal]{bottou2016optimization}
Bottou, L., Curtis, F.~E., and Nocedal, J.
\newblock Optimization methods for large-scale machine learning.
\newblock \emph{arXiv preprint arXiv:1606.04838}, 2016.

\bibitem[Bradbury et~al.(2016)Bradbury, Merity, Xiong, and
  Socher]{Bradbury2016}
Bradbury, J., Merity, S., Xiong, C., and Socher, R.
\newblock {Quasi-Recurrent Neural Networks}.
\newblock \emph{arXiv preprint arXiv:1611.01576}, 2016.

\bibitem[Cooijmans et~al.(2016)Cooijmans, Ballas, Laurent, and
  Courville]{Cooijmans2016RecurrentBN}
Cooijmans, T., Ballas, N., Laurent, C., and Courville, A.~C.
\newblock Recurrent batch normalization.
\newblock \emph{CoRR}, abs/1603.09025, 2016.

\bibitem[Duchi et~al.(2011)Duchi, Hazan, and Singer]{duchi2011adaptive}
Duchi, J., Hazan, E., and Singer, Y.
\newblock Adaptive subgradient methods for online learning and stochastic
  optimization.
\newblock \emph{Journal of Machine Learning Research}, 12\penalty0
  (Jul):\penalty0 2121--2159, 2011.

\bibitem[F{\"o}ldi{\'a}k(1991)]{Foldiak1991}
F{\"o}ldi{\'a}k, P.
\newblock Learning invariance from transformation sequences.
\newblock \emph{Neural Computation}, 3\penalty0 (2):\penalty0 194--200, 1991.

\bibitem[Gal \& Ghahramani(2016)Gal and Ghahramani]{Gal2016ATG}
Gal, Y. and Ghahramani, Z.
\newblock A theoretically grounded application of dropout in recurrent neural
  networks.
\newblock In \emph{NIPS}, 2016.

\bibitem[Goyal et~al.(2017)Goyal, Doll{\'a}r, Girshick, Noordhuis, Wesolowski,
  Kyrola, Tulloch, Jia, and He]{Goyal2017}
Goyal, P., Doll{\'a}r, P., Girshick, R., Noordhuis, P., Wesolowski, L., Kyrola,
  A., Tulloch, A., Jia, Y., and He, K.
\newblock Accurate, large minibatch sgd: Training imagenet in 1 hour.
\newblock \emph{arXiv preprint arXiv:1706.02677}, 2017.

\bibitem[Grave et~al.(2016)Grave, Joulin, and Usunier]{Grave2016}
Grave, E., Joulin, A., and Usunier, N.
\newblock Improving neural language models with a continuous cache.
\newblock \emph{arXiv preprint arXiv:1612.04426}, 2016.

\bibitem[Hardt et~al.(2015)Hardt, Recht, and Singer]{hardt2015train}
Hardt, M., Recht, B., and Singer, Y.
\newblock Train faster, generalize better: Stability of stochastic gradient
  descent.
\newblock \emph{arXiv preprint arXiv:1509.01240}, 2015.

\bibitem[Hinton(1989)]{Hinton1989}
Hinton, G.~E.
\newblock Connectionist learning procedures.
\newblock \emph{Artificial intelligence}, 40\penalty0 (1-3):\penalty0 185--234,
  1989.

\bibitem[Inan et~al.(2016)Inan, Khosravi, and Socher]{Inan2016}
Inan, H., Khosravi, K., and Socher, R.
\newblock {Tying Word Vectors and Word Classifiers: A Loss Framework for
  Language Modeling}.
\newblock \emph{arXiv preprint arXiv:1611.01462}, 2016.

\bibitem[Ioffe \& Szegedy(2015)Ioffe and Szegedy]{Ioffe2015BatchNA}
Ioffe, S. and Szegedy, C.
\newblock Batch normalization: Accelerating deep network training by reducing
  internal covariate shift.
\newblock In \emph{ICML}, 2015.

\bibitem[Jing et~al.(2016)Jing, Shen, Dub{\v{c}}ek, Peurifoy, Skirlo, Tegmark,
  and Solja{\v{c}}i{\'c}]{Jing2016}
Jing, L., Shen, Y., Dub{\v{c}}ek, T., Peurifoy, J., Skirlo, S., Tegmark, M.,
  and Solja{\v{c}}i{\'c}, M.
\newblock {Tunable Efficient Unitary Neural Networks (EUNN) and their
  application to RNN}.
\newblock \emph{arXiv preprint arXiv:1612.05231}, 2016.

\bibitem[Jonschkowski \& Brock(2015)Jonschkowski and
  Brock]{Jonschkowski2015LearningSR}
Jonschkowski, R. and Brock, O.
\newblock Learning state representations with robotic priors.
\newblock \emph{Auton. Robots}, 39:\penalty0 407--428, 2015.

\bibitem[Keskar \& Saon(2015)Keskar and Saon]{keskar2015nonmonotone}
Keskar, N. and Saon, G.
\newblock A nonmonotone learning rate strategy for sgd training of deep neural
  networks.
\newblock In \emph{Acoustics, Speech and Signal Processing (ICASSP), 2015 IEEE
  International Conference on}, pp.\  4974--4978. IEEE, 2015.

\bibitem[Kim et~al.(2016)Kim, Jernite, Sontag, and Rush]{Kim2016}
Kim, Y., Jernite, Y., Sontag, D., and Rush, A.~M.
\newblock Character-aware neural language models.
\newblock In \emph{Thirtieth AAAI Conference on Artificial Intelligence}, 2016.

\bibitem[Kingma \& Ba(2014)Kingma and Ba]{kingma2014adam}
Kingma, D. and Ba, J.
\newblock Adam: A method for stochastic optimization.
\newblock \emph{arXiv preprint arXiv:1412.6980}, 2014.

\bibitem[Koehn et~al.(2007)Koehn, Hoang, Birch, Callison-Burch, Federico,
  Bertoldi, Cowan, Shen, Moran, Zens, Dyer, Bojar, Constantin, and
  Herbst]{Koehn2007MosesOS}
Koehn, P., Hoang, H., Birch, A., Callison-Burch, C., Federico, M., Bertoldi,
  N., Cowan, B., Shen, W., Moran, C., Zens, R., Dyer, C., Bojar, O.,
  Constantin, A., and Herbst, E.
\newblock Moses: Open source toolkit for statistical machine translation.
\newblock In \emph{ACL}, 2007.

\bibitem[Krueger et~al.(2016)Krueger, Maharaj, Kram{\'a}r, Pezeshki, Ballas,
  Ke, Goyal, Bengio, Larochelle, Courville, et~al.]{Krueger2016}
Krueger, D., Maharaj, T., Kram{\'a}r, J., Pezeshki, M., Ballas, N., Ke, N.,
  Goyal, A., Bengio, Y., Larochelle, H., Courville, A., et~al.
\newblock {Zoneout: Regularizing RNNss by randomly preserving hidden
  activations}.
\newblock \emph{arXiv preprint arXiv:1606.01305}, 2016.

\bibitem[Luciw \& Schmidhuber(2012)Luciw and Schmidhuber]{Luciw2012}
Luciw, M. and Schmidhuber, J.
\newblock Low complexity proto-value function learning from sensory
  observations with incremental slow feature analysis.
\newblock \emph{Artificial Neural Networks and Machine Learning--ICANN 2012},
  pp.\  279--287, 2012.

\bibitem[Mandt et~al.(2017)Mandt, Hoffman, and Blei]{mandt2017stochastic}
Mandt, S., Hoffman, M.~D., and Blei, D.~M.
\newblock Stochastic gradient descent as approximate bayesian inference.
\newblock \emph{arXiv preprint arXiv:1704.04289}, 2017.

\bibitem[Melis et~al.(2017)Melis, Dyer, and Blunsom]{Melis2017}
Melis, G., Dyer, C., and Blunsom, P.
\newblock {On the State of the Art of Evaluation in Neural Language Models}.
\newblock \emph{arXiv preprint arXiv:1707.05589}, 2017.

\bibitem[Merity et~al.(2016)Merity, Xiong, Bradbury, and Socher]{Merity2016}
Merity, S., Xiong, C., Bradbury, J., and Socher, R.
\newblock {Pointer Sentinel Mixture Models}.
\newblock \emph{arXiv preprint arXiv:1609.07843}, 2016.

\bibitem[Merity et~al.(2017)Merity, McCann, and Socher]{smerity-revisiting}
Merity, S., McCann, B., and Socher, R.
\newblock Revisiting activation regularization for language rnns.
\newblock \emph{arXiv preprint arXiv:1708.01009}, 2017.

\bibitem[Mikolov \& Zweig(2012)Mikolov and Zweig]{Mikolov2012}
Mikolov, T. and Zweig, G.
\newblock Context dependent recurrent neural network language model.
\newblock \emph{SLT}, 12:\penalty0 234--239, 2012.

\bibitem[Mikolov et~al.(2010)Mikolov, Karafi{\'a}t, Burget, Cernock{\'y}, and
  Khudanpur]{Mikolov2010}
Mikolov, T., Karafi{\'a}t, M., Burget, L., Cernock{\'y}, J., and Khudanpur, S.
\newblock {Recurrent neural network based language model}.
\newblock In \emph{INTERSPEECH}, 2010.

\bibitem[Panageas \& Piliouras(2016)Panageas and
  Piliouras]{panageas2016gradient}
Panageas, I. and Piliouras, G.
\newblock Gradient descent converges to minimizers: The case of non-isolated
  critical points.
\newblock \emph{CoRR, abs/1605.00405}, 2016.

\bibitem[Polyak \& Juditsky(1992)Polyak and Juditsky]{polyak1992acceleration}
Polyak, B. and Juditsky, A.
\newblock Acceleration of stochastic approximation by averaging.
\newblock \emph{SIAM Journal on Control and Optimization}, 30\penalty0
  (4):\penalty0 838--855, 1992.

\bibitem[Press \& Wolf(2016)Press and Wolf]{Press2016}
Press, O. and Wolf, L.
\newblock Using the output embedding to improve language models.
\newblock \emph{arXiv preprint arXiv:1608.05859}, 2016.

\bibitem[Semeniuta et~al.(2016)Semeniuta, Severyn, and
  Barth]{Semeniuta2016RecurrentDW}
Semeniuta, S., Severyn, A., and Barth, E.
\newblock Recurrent dropout without memory loss.
\newblock In \emph{COLING}, 2016.

\bibitem[Seo et~al.(2016)Seo, Min, Farhadi, and Hajishirzi]{Seo2016}
Seo, M., Min, S., Farhadi, A., and Hajishirzi, H.
\newblock {Query-Reduction Networks for Question Answering}.
\newblock \emph{arXiv preprint arXiv:1606.04582}, 2016.

\bibitem[Srivastava et~al.(2014)Srivastava, Hinton, Krizhevsky, Sutskever, and
  Salakhutdinov]{Srivastava2014DropoutAS}
Srivastava, N., Hinton, G., Krizhevsky, A., Sutskever, I., and Salakhutdinov,
  R.
\newblock Dropout: a simple way to prevent neural networks from overfitting.
\newblock \emph{Journal of Machine Learning Research}, 15:\penalty0 1929--1958,
  2014.

\bibitem[Sutskever et~al.(2013)Sutskever, Martens, Dahl, and
  Hinton]{sutskever2013importance}
Sutskever, I., Martens, J., Dahl, G., and Hinton, G.
\newblock On the importance of initialization and momentum in deep learning.
\newblock In \emph{International conference on machine learning}, pp.\
  1139--1147, 2013.

\bibitem[Tallec \& Ollivier(2017)Tallec and Ollivier]{Tallec2017}
Tallec, C. and Ollivier, Y.
\newblock Unbiasing truncated backpropagation through time.
\newblock \emph{arXiv preprint arXiv:1705.08209}, 2017.

\bibitem[Tieleman \& Hinton(2012)Tieleman and Hinton]{tieleman2012lecture}
Tieleman, T. and Hinton, G.
\newblock Lecture 6.5-rmsprop: Divide the gradient by a running average of its
  recent magnitude.
\newblock \emph{COURSERA: Neural networks for machine learning}, 4\penalty0
  (2):\penalty0 26--31, 2012.

\bibitem[Wan et~al.(2013)Wan, Zeiler, Zhang, LeCun, and Fergus]{dropconnect}
Wan, L., Zeiler, M., Zhang, S., LeCun, Y, and Fergus, R.
\newblock Regularization of neural networks using dropconnect.
\newblock In \emph{Proceedings of the 30th international conference on machine
  learning (ICML-13)}, pp.\  1058--1066, 2013.

\bibitem[Wilson et~al.(2017)Wilson, Roelofs, Stern, Srebro, and
  Recht]{wilson2017marginal}
Wilson, A.~C, Roelofs, R., Stern, M., Srebro, N., and Recht, B.
\newblock The marginal value of adaptive gradient methods in machine learning.
\newblock \emph{arXiv preprint arXiv:1705.08292}, 2017.

\bibitem[Wisdom et~al.(2016)Wisdom, Powers, Hershey, Le~Roux, and
  Atlas]{Wisdom2016}
Wisdom, S., Powers, T., Hershey, J., Le~Roux, J., and Atlas, L.
\newblock Full-capacity unitary recurrent neural networks.
\newblock In \emph{Advances in Neural Information Processing Systems}, pp.\
  4880--4888, 2016.

\bibitem[Zaremba et~al.(2014)Zaremba, Sutskever, and Vinyals]{Zaremba2014}
Zaremba, W., Sutskever, I., and Vinyals, O.
\newblock Recurrent neural network regularization.
\newblock \emph{arXiv preprint arXiv:1409.2329}, 2014.

\bibitem[Zilly et~al.(2016)Zilly, Srivastava, Koutn{\'\i}k, and
  Schmidhuber]{Zilly2016}
Zilly, J.~G., Srivastava, R.~K., Koutn{\'\i}k, J., and Schmidhuber, J.
\newblock Recurrent highway networks.
\newblock \emph{arXiv preprint arXiv:1607.03474}, 2016.

\bibitem[Zoph \& Le(2016)Zoph and Le]{Zoph2016}
Zoph, B. and Le, Q.~V.
\newblock Neural architecture search with reinforcement learning.
\newblock \emph{arXiv preprint arXiv:1611.01578}, 2016.

\end{thebibliography}
\bibliographystyle{icml2017}

\end{document}